\documentclass{article}

\usepackage{microtype}
\usepackage{graphicx}
\usepackage{subfigure}
\usepackage{booktabs} 
\usepackage{hyperref}
\usepackage{color,soul}
\usepackage{amssymb}
\usepackage{multirow}
\usepackage{siunitx}
\usepackage{bm}
\newcommand{\todo}[1]{}
\renewcommand{\todo}[1]{{\color{red} TODO: {#1}\\}}

\usepackage[accepted]{mlsys2020}

\mlsystitlerunning{Searching for Winograd-aware Quantized Networks}

\newcommand\blfootnote[1]{%
  \begingroup
  \renewcommand\thefootnote{}\footnote{#1}%
  \addtocounter{footnote}{-1}%
  \endgroup
}

\begin{document}

\twocolumn[
\mlsystitle{Searching for Winograd-aware Quantized Networks}




\begin{mlsysauthorlist}
\mlsysauthor{Javier Fernandez-Marques}{ox}
\mlsysauthor{Paul N. Whatmough}{arm}
\mlsysauthor{Andrew Mundy}{arm}
\mlsysauthor{Matthew Mattina}{arm}
\end{mlsysauthorlist}

\mlsysaffiliation{ox}{Department of Computer Science, University of Oxford (UK)}
\mlsysaffiliation{arm}{Arm ML Research Lab}

\mlsyscorrespondingauthor{Javier Fernandez-Marques}{javier.fernandezmarques@cs.ox.ac.uk}

\mlsyskeywords{Machine Learning, Winograd, Deep Neural Networks, Neural Architecture Search, MLSys}

\vskip 0.3in

\begin{abstract}
Lightweight architectural designs of Convolutional Neural Networks (CNNs) together with quantization have paved the way for the deployment of demanding computer vision applications on mobile devices. 
Parallel to this, alternative formulations to the convolution operation such as FFT, Strassen and Winograd, have been adapted for use in CNNs offering further speedups. Winograd convolutions are the fastest known algorithm for spatially small convolutions, but exploiting their full potential comes with the burden of numerical error, rendering them unusable in quantized contexts. In this work we propose a Winograd-aware formulation of convolution layers which exposes the numerical inaccuracies introduced by the Winograd transformations to the learning of the model parameters, enabling the design of competitive quantized models without impacting model size. We also address the source of the numerical error and propose a relaxation on the form of the transformation matrices, resulting in up to 10\% higher classification accuracy on CIFAR-10. Finally, we propose wiNAS, a neural architecture search (NAS) framework that jointly optimizes a given macro-architecture for accuracy and latency leveraging Winograd-aware layers. 
A Winograd-aware ResNet-18 optimized with wiNAS for CIFAR-10 results in $2.66\times$ speedup compared to \textit{im2row}, one of the most widely used optimized convolution implementations, with no loss in accuracy. 
\end{abstract}
]



\blfootnote{This project has received funding from the European Union's Horizon 2020 research and innovation programme under grant agreement No 732204 (Bonseyes). This work is supported by the Swiss State Secretariat for Education, Research and Innovation (SERI) under contract number 16.0159. The opinions expressed and arguments employed herein do not necessarily reflect the official views of these funding bodies.\vskip 0.3in}

\printAffiliationsAndNotice{}  

\vspace{-10mm}

\section{Introduction}
\label{introduction}
The rise in popularity of deep CNNs has spawned a research effort to find lower complexity networks to increase inference efficiency. This is \textit{desirable} for inference in the cloud and becomes \textit{crucial} on mobile and IoT devices with much more constrained hardware~\cite{8364435}. Over the last few years, multiple approaches have been proposed to alleviate the compute-bound nature of convolutions~\cite{Sze_2017}. Arguably, the use of depthwise convolutions, as popularized by the family of MobileNet architectures~\cite{howard2017mobilenets,mobilenetV2,mobilenetV3}, has become the most widely embraced design choice to make lightweight networks. These layers are used in state of the art image classification networks~\cite{stamoulis2019singlepath,tan2019efficientnet}. However, beyond image classification, normal convolutions are still chosen in favour of depthwise convolutions for applications like image super-resolution \cite{RCAN,RoysonSR}, image segmentation \cite{gatedscnn} and, GANs~\cite{brock2018large}. Therefore, alternative forms of speeding up standard convolutions are required to run these applications in mobile CPUs, which often come with constrained compute and energy budgets~\cite{smiv_vlsi19}. Model quantization and the use of alternative convolution algorithms instead of direct convolution are two ways of accomplishing this task.


Lower-precision networks result in smaller model sizes, faster inference, lower energy consumption and smaller chip area~\cite{Sze_2017,memtech_dac19}. Concretely, 8-bit quantizatized models achieve comparable performance to full-precision models~\cite{Jacob_2018,whitepaper} while being ready for deployment on off-the-shelf hardware as 8-bit arithmetic is widely supported. In addition to resulting in a direct $4\times$ model size reduction, 8-bit integer-only arithmetic benefits from up to $116\times$ and $27.5\times$ chip area reduction compared to full precision additions and multiplies respectively, requiring $30\times$ and $18.5\times$ less energy~\cite{computingEnergyProblem,dnnengine_jssc18}. Because of these desirable benefits, 8-bit quantization has been widely adopted in both compute-constrained devices~\cite{liberis2019neural,chowdhery2019visual,Wang_2019_CVPR} and accelerators~\cite{whatmough2019fixynn}. 

\begin{figure}[!t]
    \centering
        \includegraphics[width=0.8\linewidth]{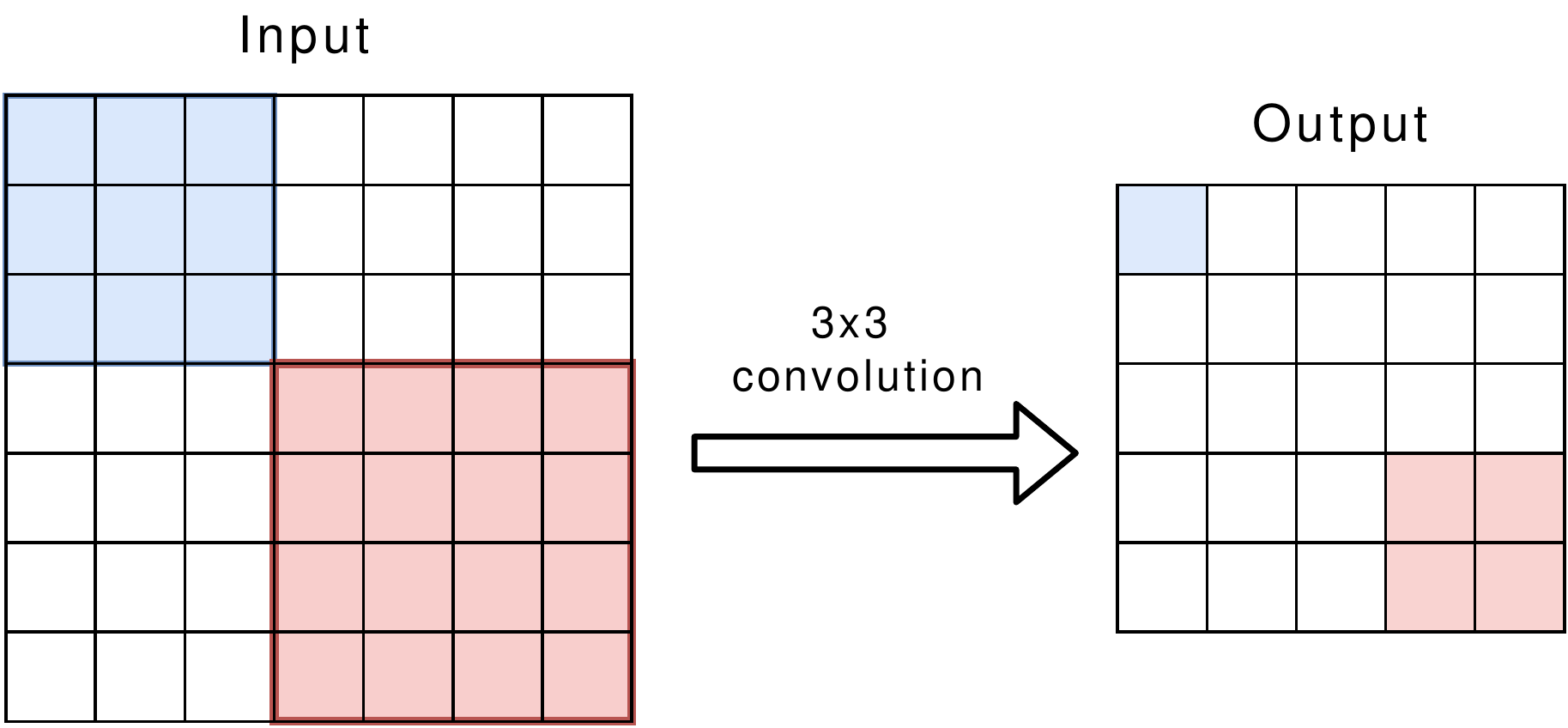}
    \caption{Standard convolutions operate on a tile (blue) defined by the filter size, generating a single output. Winograd convolutions operate on larger tiles without modifying the filter dimensions. A $3\times3$ Winograd convolution operating on a $4\times4$ tile (red) generates a $2\times2$ output. This is often expressed as $F(2\times2, 3\times3)$.}
    \label{fig:convVswinograd}
\end{figure}

Orthogonal to lightweight architectural designs and quantization, fast convolution algorithms in replacement of direct convolutions can provide further speedups. These come with their own trade-offs~\cite{Anderson_2018}, but in this work we focus on the Winograd algorithm since it is the fastest known algorithm for convolutions of the dimensions often found in CNNs. The Winograd convolution performs the bulk of the computation as a Hadamard product between weights and input in the Winograd space requiring $\mathcal{O}(n)$ operations. Unlike normal convolutions, that generate a single output per convolution, a Winograd convolution computes several outputs simultaneously. This property makes Winograd convolutions minimal in the number of \textit{general multiplications}\footnote{\textit{General multiplications} is a term commonly used in Winograd jargon referring to element-wise or Hadamard product stage.}\cite{winogradGeneralMultiplications}. Normal convolutions operate on tile sizes matching the width and height of the filter, on the other hand, Winograd convolutions can operate on larger tiles. This is illustrated in Figure \ref{fig:convVswinograd}. In this way, while normal convolutions would require 8.1K multiplications to densely convolve a $32\times32$ input with a $3\times3$ filter, Winograd convolutions operating on a $4\times4$ tile require only 3.6K. The speedups Winograd convolutions offer increase with tile size. 
However, exploiting these speedups exposes a problem inherent in current Winograd convolution implementations: numerical error. This error, which grows exponentially with tile size, is the primary reason why Winograd is generally only deployed with 32-bit floating point and for comparatively small tile sizes, rarely larger than $6\times6$. In practice, an architecture with standard convolutional layers would first be trained before replacing standard convolution with Winograd convolution for deployment.

In this paper, we focus on alleviating the problem of numerical error that arises when 
using Winograd convolutions in quantized neural networks. Achieving this ultimately enables us to combine the speedups of Winograd with those that reduced precision arithmetic is known to offer, among other benefits in terms of energy and area. To this end, we present an end-to-end training pipeline that exposes the numerical inaccuracies introduced by Winograd to the learning of the model parameters. 
We also address the source of the numerical error and propose a relaxation on the form of the transformation matrices used in the Winograd convolution algorithm. We achieve this by adding these matrices to the set of \textit{learnable} parameters in a layer, after initializing them via Cook-Toom~\cite{Toom1963TheCO}. 
Finally, we describe wiNAS, a Winograd-aware Neural Architecture Search framework which leverages Winograd-aware layers and latency measurements on Arm Cortex-A73 and A53 cores, to jointly optimize for high accuracy and low latency. Our framework transforms a given macro-architecture by replacing each convolution with either \textit{im2row} or Winograd convolutions of different tile sizes.

The contributions of this work are summarized below:

\begin{itemize}
    \item We show that Winograd-aware networks enable Winograd convolutions in quantized networks, including 8-bit networks with little accuracy drop. To the best of our knowledge, this is the first time this has been empirically demonstrated. 
    \item We demonstrate that learning the Winograd transforms, as opposed to keeping these fixed, results in better network accuracy -- up to 10\% improvement when using $6\times6$ and $8\times8$ tiles in 8-bits CNNs with $3\times3$ filters. This improvement is more pronounced with larger $5\times5$ filters. 
    \item We present wiNAS as a tool that can find Winograd-aware networks jointly optimised for both high accuracy and low latency given a real hardware model. 
    
\end{itemize}

\section{Related Work}
\label{relatedwork}

Convolutions have become the \textit{de facto} spatial feature extractor in neural networks. 
As a result, a number of approaches have emerged to reduce the computational costs of using this operator.


\paragraph{Compact CNN Architectures.}  
These include alternative formulations to the dense convolutional layer, such as \textit{bottleneck} layers~\cite{He_2016} that perform the $3\times3$ convolutions in a lower-dimensional space, or the \textit{depth-wise} convolutional layers~\cite{howard2017mobilenets} which replace the standard convolutions with a channel-wise convolution followed by a $1\times1$ point-wise convolution. More recently, \citet{chen2019drop} proposed a compact multi-resolution convolutional block that reduces spatial redundancy of low frequencies resulting in faster inference, memory savings and slightly higher accuracy. This reinforces the proposition that current networks rely more on texture than shape for image/object discrimination~\cite{brendel2018approximating,texture}. In this work, instead of presenting a new architecture, we propose an optimization for an existing known-good architecture to speed up inference. Our optimization can be applied to existing pre-trained models without the need for end-to-end training. 

\paragraph{Quantization.} The most extreme form of quantization is binary networks~\cite{binaryConnect,towards_accurate_binary_cnn,Xiang2017}, which replace convolutions with bit-shifts resulting in $58\times$ inference speed-ups~\cite{xnor}. Ternary and 2-bit models~\cite{li2016ternary,ternary, gong2019differentiable} achieve higher accuracies while alleviating some the challenges of training binary networks~\cite{alizadeh2018a}. However, it is 8-bit quantization~\cite{Jacob_2018,whitepaper,Wang_2019_CVPR} that has achieved high popularity due to its balance between accuracy, model size reduction and inference speedup. Newer data formats, such as Posit~\cite{Carmichael_2019} aim to close the accuracy gap between INT8 and FP32 networks, however hardware supporting it is unavailable. For training, BFLOAT16~\cite{kalamkar2019study} has been validated as an alternative to FP32, enabling faster training. In this work, we adopt INT8 and INT16 uniform quantization during training and study how lowering precision impacts on the lossy nature of Winograd convolutions.

\paragraph{Fast Convolution Algorithms.}
Alternative formulations of the convolution operation such as: the use of FFTs, which replace convolution with its multiplication-only counterpart in the frequency domain resulting in faster inference~\cite{mathieu2013fast,FFT1} and training~\cite{Highlander_2015}; the Strassen algorithm~\cite{Strassen:1969:GEO:2722431.2722798}, which when applied to convolutions ~\cite{strassenXiao,strassenTschannen} significantly reduces the number of multiplications at the cost of more additions; or the Winograd algorithm~\cite{winograd}, which replaces convolutions with a set of matrix transformations and point-wise multiplications and, results in significantly faster inference stages~\cite{Lavin_2016}.

\paragraph{Winograd Convolution.} The Winograd algorithm for fast convolution was first applied to CNNs by~\citet{Lavin_2016}, showing $2.2\times$ speedup compared to cuDNN~\cite{cuDNN} on a VGG~\cite{vgg} network, with no loss in accuracy on small $4\times4$ tiles, batch 1. 
However, exploiting Winograd convolutions on larger input tiles is challenging due to numerical instability. 
In response to this limitation, ~\citet{Barbara} showed that the error introduced by the Winograd algorithm grows \textit{at least exponentially} with tile size, which can be partially alleviated by choosing better polynomial points for constructing the transformation matrices via Cook-Toom~\cite{Toom1963TheCO}. 
An alternative formulation using trimmed Vandermonde matrices was described by~\citet{winogradICLRW17}.
More recently, several works studying the suitability of Winograd convolutions in memory and compute constrained setups have been proposed. 
These include: the use of integer arithmetic for complex Winograd convolutions~\cite{MengAndBrothers}; a general formulation for the Winograd algorithm~\cite{barabasz2019winograd} that shows promising results in FP16 and BFLOAT16 when using higher degree polynomials; an efficient region-wise multi-channel implementation of Winograd convolutions using General Matrix Multiplications (GEMMs)~\cite{maji2019efficient} that achieves $4\times$ speedups on Arm Cortex-A CPUs; and, a technique~\cite{liu2018efficient} that enables up to 90\% sparsity in the Hadamard product stage of the Winograd algorithm, effectively reducing by $10\times$ the number of multiplications with no accuracy loss in FP32 models. 
Our work fills the gap of using Winograd convolutions in quantized neural networks, enabling even faster convolutions in current off-the-shelf hardware, such as mobile CPUs.  


\paragraph{Neural Architecture Search.} Automating the process of designing neural network architectures has drawn considerable attention. Early attempts relied on reinforcement learning~\cite{NAS,brock2018smash,amoebanet,MNasNet} or Bayesian optimization~\cite{lobato_nips16,sparse_neurips19} and required thousands of GPU hours to converge due to their computationally expensive and exhaustive search stages. Other works opted instead for a gradient-based search by framing the problem as a single over-parameterized network where all \textit{candidate operations} at a particular node (e.g. a layer) are taken into consideration. The main aspect differentiating gradient-based NAS approaches is the way the output of a layer combines the contribution of each candidate operation. While \citet{bender18a} defines it as the sum and DARTS~\cite{liu2018darts} as a weighted sum, ProxylessNAS~\cite{cai2018proxylessnas} relies on path-level binarization, 
making it possible to perform the search on the entire architecture directly using a single GPU. 
In addition to architecture discovery, NAS has also been successfully used for automated network pruning~\cite{He_2018} and quantization~\cite{Wang_2019_CVPR}. Our work leverages NAS to find the optimal convolution algorithm (i.e. \textit{im2row} or different Winograd implementations) for each layer in the model while preserving the overall network macro-architecture and model size.

\section{Winograd-Aware Networks}
\label{winogradawarenets}

This section introduces Winograd convolutions and their trade-offs in terms of compute, memory and accuracy. Then, we present the Winograd-aware layers used in our networks.

\subsection{Winograd implementation trade-offs}

The Winograd algorithm for convolutions using linear polynomials guarantees to use the minimum number of element-wise multiplications to compute $m\times m$ outputs using an $r\times r$ filter. \citet{Lavin_2016} refer to this minimal algorithm as $F(m\times m, r\times r)$ and present its matrix form:

\begin{equation}
    Y=A^{\top}\Big[[GgG^{\top}]\odot [B^{\top}dB]\Big]A
    \label{eq:winogradmatrix}
\end{equation}{}

\noindent
where $G$, $B$ and $A$ are transformation matrices applied to the filter $g$, input and output respectively and $\odot$ is the Hadamard or element-wise multiplication. 

These transformation matrices are commonly constructed\footnote{See Section 5.2 in \citet{blahut_2010} for a step-by-step example.} as described in the Cook-Toom algorithm which requires choosing a set of so-called \textit{polynomial points} from $\mathbb{R}^{2}$. This choice is not trivial, but for small Winograd kernels e.g. $F(2\times2,3\times3)$ or $F(4\times4,3\times3)$, there is a common consensus. While a standard convolution using a $r\times r$ filter $g$ would operate on a $r\times r$ input tile, a Winograd convolution expressed as Eq.~\ref{eq:winogradmatrix} expects an input patch $d$ with dimensions $(m+r-1)\times(m+r-1)$. The key difference is that while the standard convolution would generate a $1\times1$ output, the Winograd convolution would compute a $m\times m$ output. In this way, a standard $3\times3$ convolution requires 9 mult. per output (\textit{mpo}), $F(2\times2,3\times3)$ and $F(4\times4,3\times3)$ require 4 \textit{mpo} and 2.25 \textit{mpo} respectively. Theoretically, these savings grow as we increase the tile or filter sizes. For the remaining of this work and, unless stated otherwise, we will be considering $3\times3$ filters and therefore refer to $F(2\times2,3\times3)$ as $F2$, $F(4\times4,3\times3)$ as $F4$, and so on. 

The challenges associated with the use of Winograd convolutions span three dimensions:

\textbf{Compute.} Winograd convolutions require the transformation of both tile $d$ and filter $g$ to the Winograd domain. The cost of these transformations grows with $m$, and can represent a significant portion of the total computation of up to 75\% (Sec. \ref{subsec:winogradStudy}). This suggests that Winograd offers little to no speedup in layers with few filters. The cost of $GgG^{\top}$ is often ignored as it is amortized across inferences. 
    
\textbf{Memory.}  In Eq.\ref{eq:winogradmatrix}, $GgG^\top$ transforms the filter $g$ to the Winograd domain, matching the dimensions of the input tile $d$. This results in an increase of run-time memory associated with the weights: $1.78\times$ and $4\times$ for $F2$ and $F4$ respectively. This is especially undesirable on memory-constrained devices such as microcontrollers.
    
\textbf{Numerical Error.} Small $F2$ and $F4$ perform well in single and double precision (FP32/64) networks and are available in production-ready libraries such as NVIDIA cuDNN~\cite{cuDNN} and Arm Compute Library~\cite{armComputLibrary}. Because these introduce only marginal numerical error, a network can first be trained using conventional convolutions before replacing appropriate layers with Winograd, without impacting accuracy. 
However, attempting this with larger Winograd tiles, or in combination with quantization, results significant accuracy loss. The root of the problem\footnote{We refer the interested reader to \citet{Barbara} for an analysis on the nature of the errors in Winograd convolutions.} is the increasing numerical range in $G$, $B$ and $A$ as $d$ increases. As a consequence, the multiple matrix multiplications in Eq.\ref{eq:winogradmatrix} contribute considerable error, ultimately reducing accuracy. This problem is exacerbated in networks using quantized weights and activations, where the range and precision of values is reduced. We show these limitations in Table~\ref{tab:rawWinograd}. The numerical error is, to a large extent, the main limiting factor for adopting large-tiled Winograd and for adopting Winograd convolutions in general for reduced precision networks. 

\begin{table}[t]
    \small
    \centering
    \begin{tabular}{l S S S}
        \toprule
              & \multicolumn{3}{c}{ResNet-18 Accuracy} \\
        \cmidrule{2-4}
        Convolution method &  {32-bit} & {16-bit} & {8-bit} \\
        \midrule
        Direct  & 93.16 & 93.60 & 93.22 \\
        {Winograd {$F2$}}   & 93.16  & 93.48 & 93.21 \\
        {Winograd {$F4$}}   & 93.14 & 19.25  & 17.36 \\
        {Winograd {$F6$}}  & 93.11  & 11.41 & 10.95 \\
        \bottomrule
    \end{tabular}
    \caption{Replacing the convolutional layers in pre-trained ResNet-18 models on CIFAR-10 with $F2$, $F4$ and $F6$. This works well in full precision, but accuracy drops drastically with quantization for configurations beyond $F2$. Note that prior to evaluating the quantized configurations we performed a \textit{warmup} of all the moving averages involved in Eq.\ref{eq:winogradmatrix} using the training set but without modifying the weights. Without this relaxation (which requires a Winograd-aware pipeline as in Fig. \ref{fig:pipeline}), $F2$ would be unusable.}
    \label{tab:rawWinograd}
\end{table}

In this work we focus on minimizing the numerical errors that arise when using the Winograd algorithm in quantized networks. Our approach does not aggravate the compute and memory challenges previously mentioned. Instead, it indirectly alleviates these by making use of quantization.

\subsection{A Winograd-aware training pipeline}

\begin{figure*}[!t]
    \centering
        \includegraphics[width=0.95\linewidth]{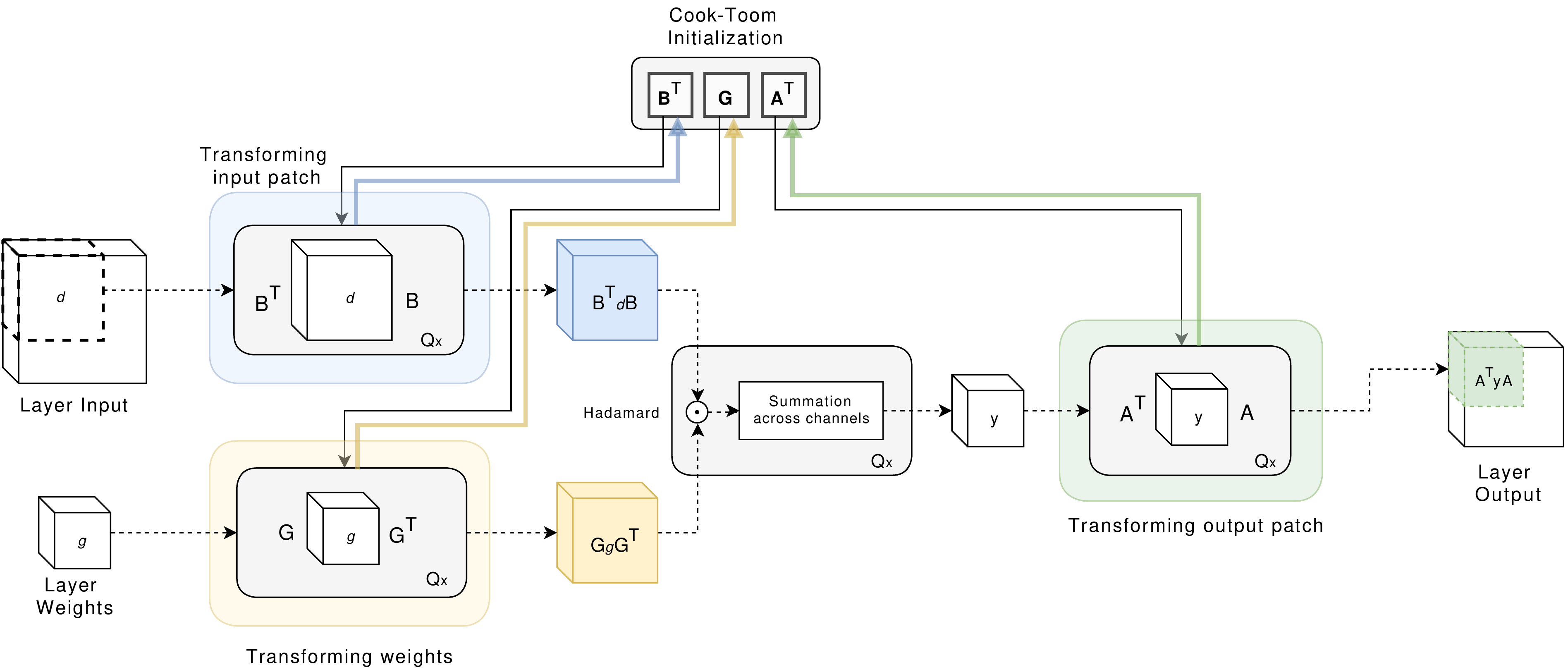}

    \caption{Forward pass of Winograd-aware layers. Transformation matrices $G$, $B^\top$ and $A^\top$ are constructed via Cook-Toom. If these are included in the set of model parameters, they would be updated with every batch via back-progation (this is represented with the coloured arrows going back to matrices $G$, $B^\top$ and $A^\top$, carrying the gradients to update the values of each transform). In its default configuration, each intermediate output throughout the pipeline quantized to the same level as the input and weights, this is represented by $Qx$. }
    \label{fig:pipeline}
\end{figure*}

Neural networks have proven to be resilient to all kinds of approximations, e.g. pruning and quantization. When applying these techniques, consistently better models are generated if these approximations are present during training. In other words, when the training is \textit{aware} of quantization
, or when training is \textit{aware} of pruning.

Following this intuition, we propose an end-to-end Winograd-aware pipeline as shown in Figure~\ref{fig:pipeline}. In the forward pass we apply Eq.\ref{eq:winogradmatrix} to each patch of the activations from the previous layer. We can apply standard back-propagation, since Eq.\ref{eq:winogradmatrix} is only a collection of matrix-matrix multiplications. This implementation allows us to:

\begin{itemize}
    \item \textbf{Learn better filters.} Building an explicit implementation of each of the stages involved in the Winograd transform exposes the numerical errors introduced in Eq.\ref{eq:winogradmatrix} to the learning of the filters. This prevents the accuracy drops shown in Table~\ref{tab:rawWinograd}.
    \item \textbf{Learn the transforms.} Traditionally, matrices $G$, $B^\top$ and $A^\top$ are fixed. Instead, we can treat them as another set of \textit{learnable} parameters in the layer. This relaxation leads to much improved performance in quantized networks while still maintaining the overall structure of the Winograd convolution algorithm and its speedups.
    \item \textbf{Quantization diversity.} Unlike standard convolution, which does not require intermediate computation, Winograd convolution requires at least four of them for $GgG^\top$, $B^\top dB$, the Hadamard product and the output transformation. Each of these can be quantized to a different number of bits depending on the bit-width of the input, that of the weights, and the overall complexity of the problem the network is designed to solve.
\end{itemize}

\section{Searching for Winograd-Aware Networks}
\label{sesarchforwinogradawarenets}
Simultaneously maximizing accuracy and minimizing latency with Winograd convolution isn't trivial. The reason for this is that large tiles result in low latency, but come at the cost of higher numerical error. This presents a good opportunity to jointly optimize network accuracy and latency. 

To this end, we implement a NAS-based approach that automatically transforms an existing architecture into a Winograd-aware version. We perform NAS at the micro-architecture level by selecting from different convolution algorithms for each layer, but without modifying the network's macro-architecture (e.g. number or order of layers, hyper-parameters, etc). Keeping the macro-architecture fixed allows us to fairly compare the standard model to its Winograd-aware counterpart in terms of latency and accuracy. We call our framework wiNAS. 

\subsection{Winograd-aware NAS pipeline}

Introducing latency measurements into the optimization objective requires knowing the shape of the input tensor, i.e. the activations from the previous layer, at each layer of the network. 
We design wiNAS as a variation of ProxylessNAS~\cite{cai2018proxylessnas}, leveraging path sampling while performing the search. This technique, enables the allocation of the entire network on a single GPU by evaluating no more than two candidate operations at each layer per batch. 



Similarly to ProxylessNAS, wiNAS formulates the search as a two-stage process, alternating the update of model parameters (the weights), where the loss is defined as

\begin{equation}
    L_{weights} = Loss_{CE} + \lambda_{0} \left \| w \right \|^{2}_{2}
\end{equation}{}

\noindent
and the update of architecture parameters (the weight assigned to each operation on a given layer), where the loss introduces the latency metrics is defined as

\begin{equation}
    L_{arch} = Loss_{CE} + \lambda_{1} \left \| a \right \|^{2}_{2} + \lambda_{2}E\{latency\} 
    \label{eq:arch_search_loss}
\end{equation}{}

\noindent
where $a$ are the architecture parameters and $\lambda_{2}$ controls the impact of latency in the loss. The expected latency, $E\{latency\}$, for a given layer is the weighted combination of the latency estimate of each candidate operation with their respective probability of being sampled. Intuitively, searching for Winograd convolutions with high $\lambda_{2}$ would result in faster models, potentially at the detriment of accuracy.

\begin{figure}[!t]
    \centering
        \includegraphics[width=0.95\linewidth]{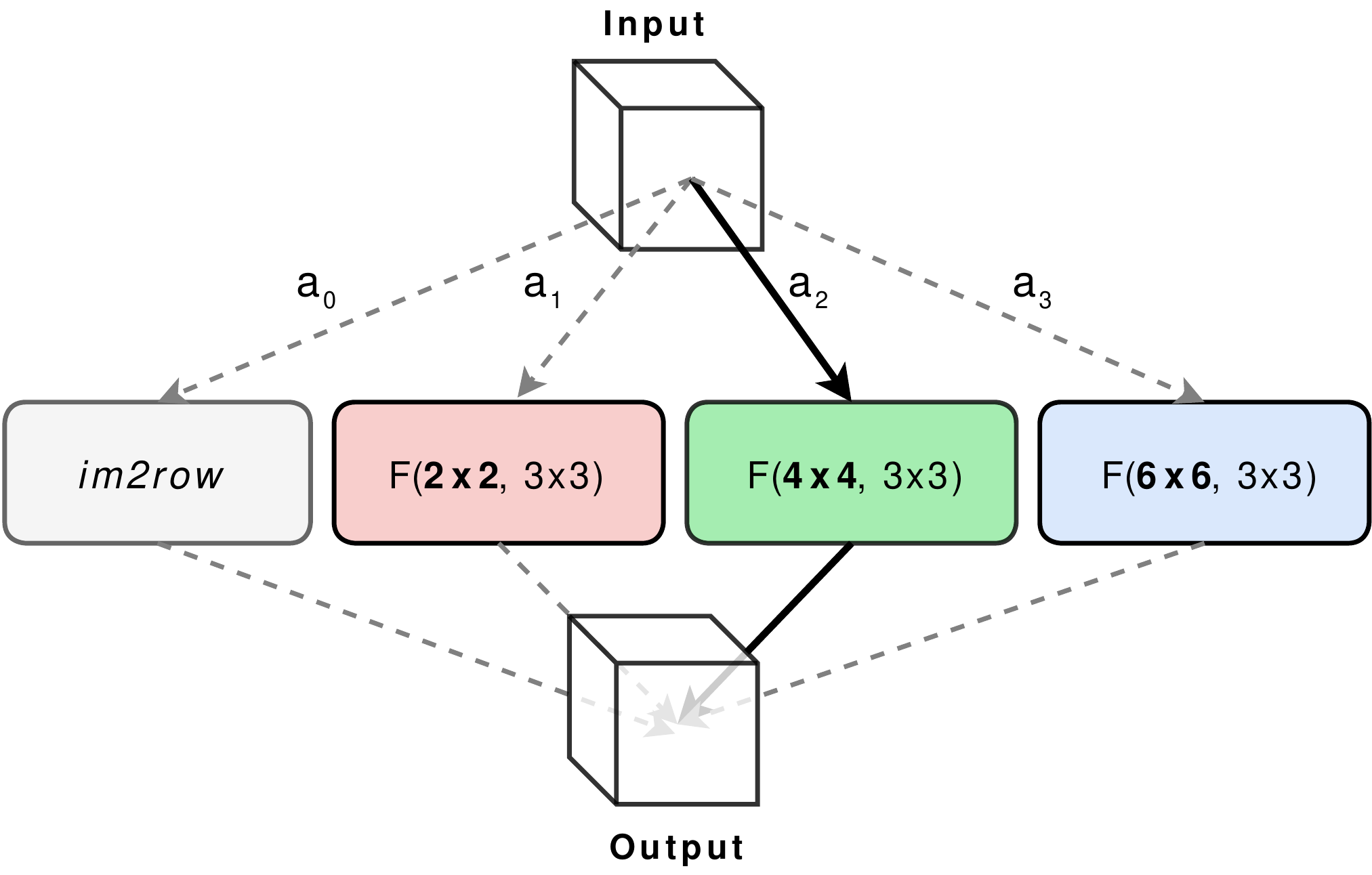}
    \caption{With wiNAS, each $3\times3$ convolution in a given architecture is implemented with either \textit{im2row} or with Winograd convolutions varying tile size. While the former is lossless and faster than direct convolution, Winograd offers lower latencies but introduce numerical instability that could ultimately impact in accuracy.}
    \label{fig:pathSampling}
\end{figure}

Unlike ProxylessNAS, wiNAS focuses on simply selecting the optimal convolution algorithm for each of the $3\times3$ convolutional layers. Therefore, the set of candidate operations for a given \textit{conv2d} layer contains \textit{im2row} and Winograd-aware layers in their $F2$, $F4$ and $F6$ configurations. This search space is illustrated in Figure \ref{fig:pathSampling}. Each candidate operation comes with its respective latency, which is a function of the output dimensions and quantization level.

\section{Experimental Setup}
\label{experimentalsetup}

We conduct various experiments grouped in three categories. In this section we describe each experiment we conducted. 
We used PyTorch~\cite{pytorch} for training and Arm Compute Library for deployment. 

\subsection{Vanilla Winograd-aware networks}
\label{experimentalsetup_WA_networks}

We begin our study of Winograd-aware networks by performing an extensive evaluation on the ResNet-18~\cite{He_2016} architecture using the CIFAR-10~\cite{krizhevsky2009learning} dataset. In this experiment we train the network end-to-end using standard convolutions, and $F2$, $F4$ and $F6$ Winograd convolutions. For each experiment with Winograd, all layers in the network use the same tile size, except the last two residual blocks which are kept fixed to $F2$.
 The input convolutional layer uses normal convolutions. We run the experiments for FP32, INT16, INT10 and INT8 quantized networks, where both weights and activations are uniformly quantized (including all the intermediate outputs shown in Figure \ref{fig:pipeline}). We follow the per-layer symmetric quantization as described in \citet{whitepaper}.
We repeated each experiment while enabling the Winograd transforms $G$, $B^\top$ and $A^\top$ to be learnt, which we denote using the additional suffix \textit{-flex}. 

Winograd-aware layers do not require an over-parameterized model to perform well. We also varied the model size by using a width-multiplier, as used by the MobileNets family, ranging from 0.125 to 1.0, meaning that when the multiplier is 1.0 the network is the full ResNet-18. This leads to models ranging between 215K and 11M parameters. Winograd-aware layers with learnable transformations marginally increase ($<0.1\%$) the model size, since the transforms themselves need to be saved for model deployment. 
We repeated the experiment for CIFAR-100~\cite{krizhevsky2009learning}, but without varying the depth-multiplier. CIFAR-100 is considerably more challenging that CIFAR-10, as it is comprised of 100 classes with only 600 images per class.


Additionally, we use an INT8 LeNet~\cite{LeNet}, trained on the MNIST dataset, to evaluate the suitability of Winograd-aware layers with learnable transforms for $5\times5$ filters.  This is a more challenging case than $3\times3$ filters, because a larger tile tile is required (defined by $F(m\times m, r\times r)$), with larger transformation matrices which require the choice of more good polynomial points.

For experiments on ResNet-18, we replace $2\times2$-stride convolution layers with a $2\times2$ max-pooling layer followed by a dense $3\times3$ convolution layer. Altering the network in this way is necessary since there is no known equivalent for strided Winograd convolutions, which remains an open research question. This is a common strategy when evaluating Winograd~\cite{liu2018efficient,choi2018compression}. We also modified the number of output channels of the input layer from 64 to 32. We did this to reduce the memory peak during training. We use the Adam~\cite{kingma:adam} optimizer 
and train for 120 epochs. Both CIFAR-10/100 use the same ResNet-18 architecture, differing only in the number of outputs of the fully connected layer. 
Results for other architectures are shown in ~\ref{otherArchitectures}.

\subsection{wiNAS: Winograd-aware NAS}


To evaluate wiNAS, we define two different sets of \textit{candidate operations}. These spaces are: wiNAS\textsubscript{WA} and wiNAS\textsubscript{WA-Q}, both allowing each $3\times3$ convolutional layer to be implemented with either \textit{im2row} or each of the Winograd configurations, $F2$, $F4$ or $F6$. The former uses a fixed bit-width for all elements in the architecture, while the latter introduces in the search space candidates of each operation quantized to FP32, INT16 and INT8. 


The hyperparameters used for wiNAS are as follows: for the learning of model parameters we use mini-batch SGD with Nesterov momentum~\cite{nesterov}. In the stage where we update the architecture parameters we use instead Adam with the first momentum scaling, $\beta_{1}$, set to zero, so the optimizer only updates paths that have been sampled. For both stages we use Cosine Annealing~\cite{cosineAnealing} scheduling and a batch size of 64. We perform the search for 100 epochs in each search space at different $\lambda_{2}$ values ranging from 0.1 to 1e-3. 
Once the search is completed, we trained the architecture end-to-end with the same hyperparameters as the rest of winograd-aware networks.

\subsection{Winograd convolutions on mobile CPUs}
\label{benchmarks}

For our study, we chose Arm A73 and A53 cores on a Huawei HiKey 960 development board with the big.LITTLE CPU architecture. These cores are good candidates for validating the speedups that are achievable with Winograd convolutions in today's off-the-shelf 
mobile hardware. 

\begin{table}[h]
    \centering
    \small
    \begin{tabular}{l c c c }
        \toprule
        CPU  & Clock & L1 & L2  \\
        \midrule
        A73 & 2.4 GHz & 64 KB & 2048 KB   \\
        A53 & 1.8 GHz & 32 KB  & 512 KB \\
        \bottomrule
    \end{tabular}
    \caption{Key hardware specifications for the high-performance Cortex-A73 and the high-efficiency Cortex-A53 cores found on a HiKey 960 development board. 
    }
    \label{tab:CoreSpecs}
\end{table}

While both A73 and A53 are implemented as 16nm quad-core CPUs, the former is a high-performance processor and the latter implements a high-efficiency processor. In Table \ref{tab:CoreSpecs} we summarise the main differences between these CPUs. The memory bandwidth would be the primary factor that ultimately sets the upper limit to the speedup achievable by Winograd since it requires operating in larger tiles than direct convolution algorithms such as \textit{im2row} or \textit{im2col}. 

In our study, we measured the time taken for $3\times3$ convolutions using \textit{im2row}, \textit{im2col} and each of the Winograd configurations ($F2$, $F4$, $F6$) when varying output width/height (from $112\times112$ down to $2\times2$) and $inCh\rightarrow outCh$ (from $3\rightarrow32$ to $512\rightarrow512$). We performed the benchmark in controlled conditions and in single thread mode. Each combination was run five times with five seconds delay in between to prevent thermal throttling. We implemented Winograd convolutions using GEMMs (\citet{maji2019efficient}), and performed the same experiment separately on A73 and A53 for both FP32 and INT8. INT16 measurements are not currently supported in Arm Compute Library.
\section{Experimental Results}
\label{experimentalresults}

The results of this work are arranged as three subsections. First, we show that winograd-aware networks can achieve high accuracy. Second, we present the results from our dense benchmark for winograd convolutions on mobile CPUs. Third, we show that wiNAS can jointly optimize a given macro-architecture for accuracy and latency.

\begin{figure*}[!t]
    \centering
    \includegraphics[width=\linewidth]{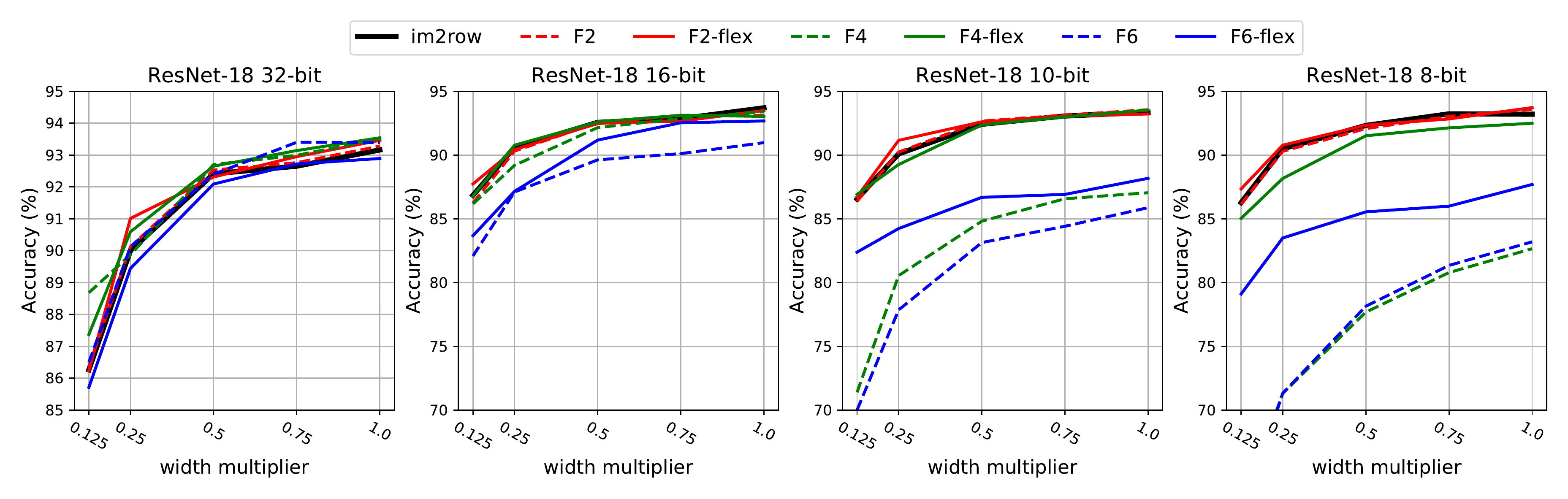}
    \vspace{-9mm}
    \caption{Performance of a winograd-aware ResNet-18 at different bit-widths and trained with different Winograd configurations. We show how winograd-aware layers scale with network's width. We can observe that in quantized networks, models that learn the Winograd transforms (\textit{-flex} configurations), strictly outperforms those models that keep them fixed with the values obtained via Cook-Toom.}
    \label{fig:flexvsstatic}
    \vspace{-5mm}
\end{figure*}

\begin{figure}[t]
    \begin{center}
    \includegraphics[width=\linewidth]{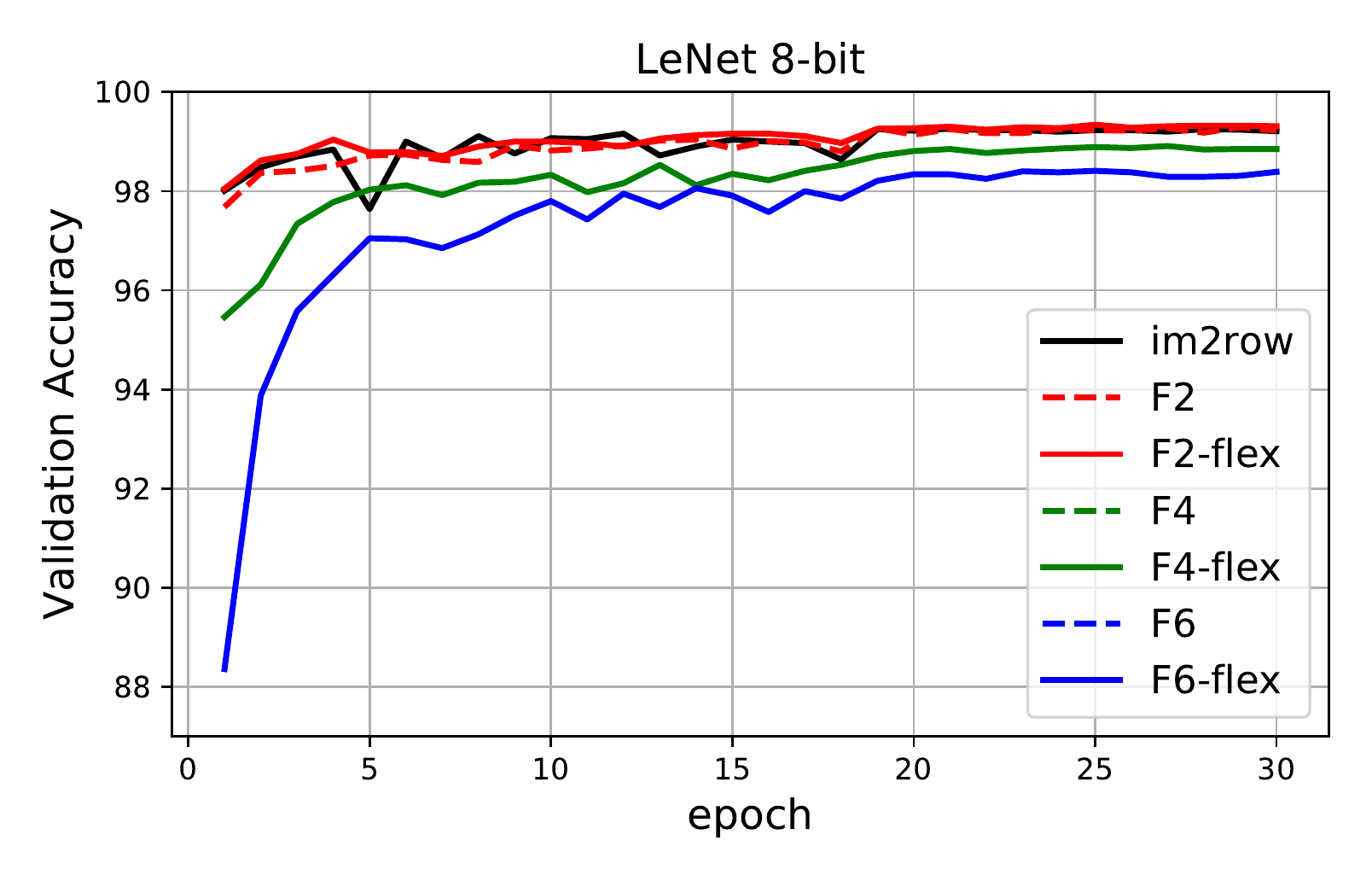}
    \end{center}
    \vspace{-7mm}
    \caption{Performance of INT8 LeNet on MNIST using standard convolutions (\textit{im2row}) or winograd-aware layers. Letting the transformations to evolve during training (\textit{-flex}) always results in better models. 
    F4 and F6 configurations (not shown) reach an accuracy of $73\%$ and $51\%$, respectively. All configurations reach $99.25\% \pm 0.1\%$ in full precision.}
    \label{fig:leNet}
    \vspace{-2mm}
\end{figure}

\subsection{Vanilla Winograd-aware networks}

Figure~\ref{fig:flexvsstatic} (left) shows  Winograd-aware networks in FP32 perform as well as direct convolutions, with both fixed and learned (\textit{-flex}) transformation matrices. 
With quantization (all other plots), winograd-aware layers are essential to enable the use of fast Winograd convolutions. This is not possible if switching to Winograd convolutions after training, as is commonly done in practice (see Table \ref{tab:rawWinograd}). 

Furthermore, we show that learning the Winograd transforms (\textit{-flex}) 
results in $10\%$ and $5\%$ better accuracies for $F4$ and $F6$ in INT8 scenarios. We argue that enabling this relaxation helps in easing the numerical instability inherent to Winograd convolutions, which is further exacerbated by quantization. The accuracy of Winograd-aware models scales linearly with network width, suggesting that these can be exploited in conjunction with architecture compression techniques such as channel pruning.

Results from LeNet ($5\times5$ filters), provides further evidence that larger tiles result in higher numerical error. In Figure~\ref{fig:leNet}, we show that even in relatively small datasets like MNIST, keeping the transformations $G$, $B^\top$ and $A^\top$ fixed, leads to poor results as the output tile size is increased. This difference is almost 47\% in the case of $F(6\times6,5\times5)$ layers, which uses $10\times10$ tiles. 

\begin{figure}[t]
    \begin{center}
    \includegraphics[width=\linewidth]{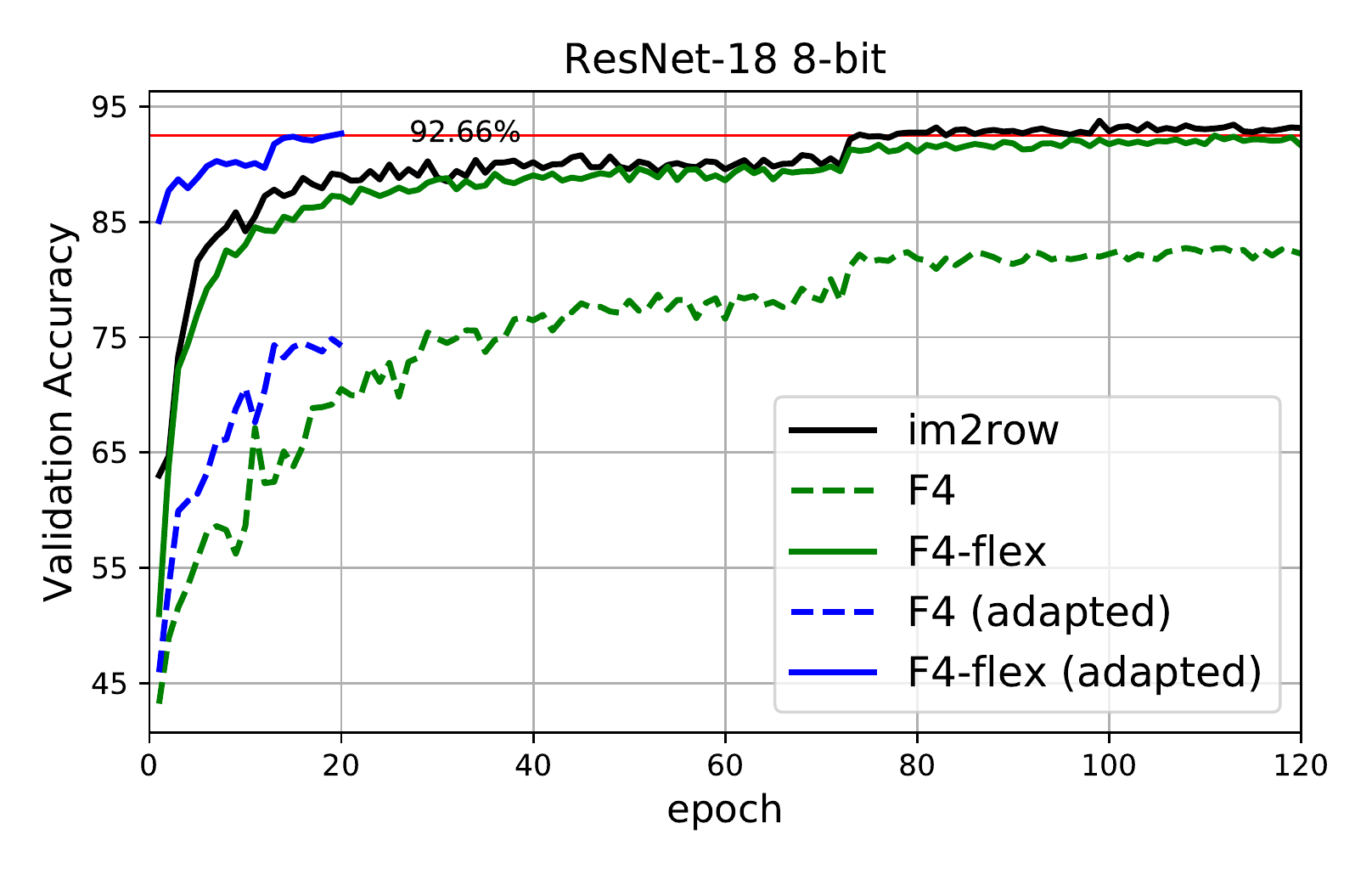}
    \end{center}
    \vspace{-7mm}
    \caption{Transforming a standard model (trained with default convolutions) to its Winograd-aware counterpart can be done in very few epochs of retraining. We found this technique works best if the Winograd transformations are learnt during retraining (\textit{-flex}), otherwise adaptation becomes much more challenging.}
    \label{fig:adapt}
    \vspace{-2mm}
\end{figure}

Winograd-aware layers do not structurally modify the network architecture, since Winograd is just an algorithm to perform convolution. We demonstrate it is possible to transform a pre-trained model with standard convolution into its Winograd-aware counterpart within a few epochs. Concretely, in Figure~\ref{fig:adapt} we show that an INT8 ResNet-18 $F4$ can be adapted from a model of the same network that was trained end-to-end with standard convolutions in 20 epochs of retraining. This represents a $2.8\times$ training time reduction for Winograd-aware models. This is only possible when allowing the transformation matrices to evolve during training. Adapting FP32 models can be done in a single epoch. 

We believe both $F4$ and $F6$ performance could be raised with alternative quantization implementations, closing the accuracy gap with $F2$ and direct convolutions. 

\subsection{Impact of Winograd on Latency}

\label{subsec:winogradStudy}

\begin{figure*}[!t]
    \centering
    \includegraphics[width=\linewidth]{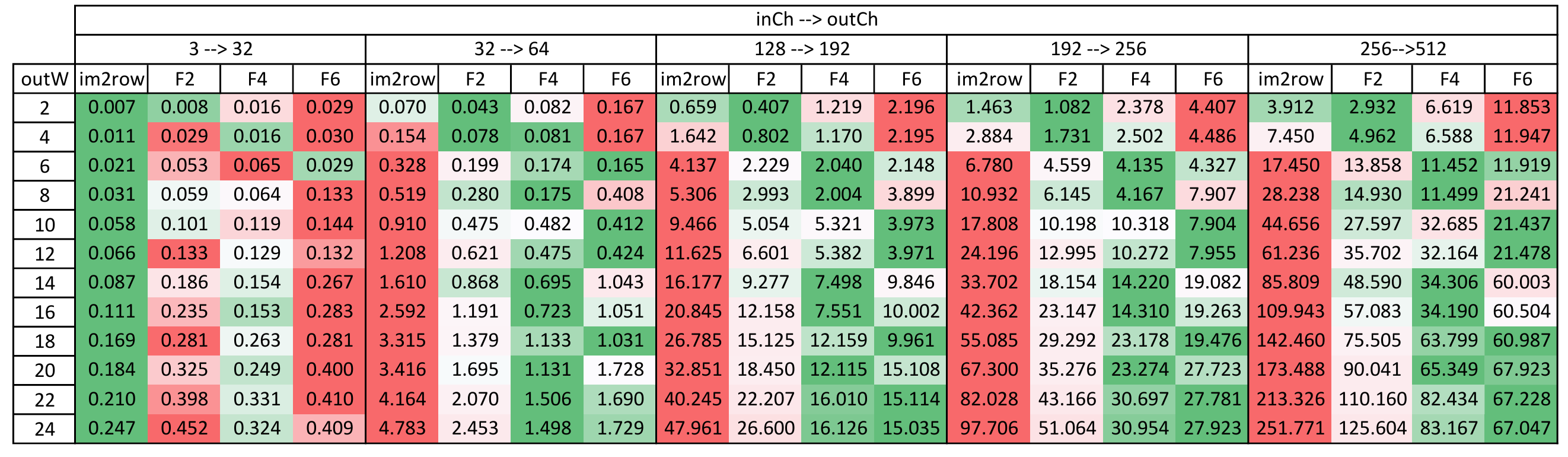}
    \vspace{-7mm}
    \caption{Latencies (in milliseconds) of convolving increasingly larger input tensors in the width/height dimensions (y axis) and in depth (x axis). We compare the time needed for \textit{im2row} and each of the Winograd configurations with 32-bit arithmetic on a Cortex-A73. We show that (1) \textit{im2row} is the consistently the optimal algorithm for the input layer to a network, (2) the choice between $F2$, $F4$ and $F6$ should be done based on the output's width/height and, (3) this choice should not generally be altered based on $inCh\rightarrow outCh$.}
    \label{fig:latency}
\end{figure*}

The speedups associated to with use of Winograd convolutions often only account for the point-wise stage while assuming negligible costs for the input, weights and output transformations. Furthermore, these also assume that the larger the input patch, \textit{d}, the larger the speedup compared to normal convolutions. However, although these assumptions are true for large tensors, they are not necessarily true when working with tensors of the sizes often found in CNNs for image classification or object detection.



\begin{figure}[t]
    \begin{center}
    \includegraphics[width=\linewidth]{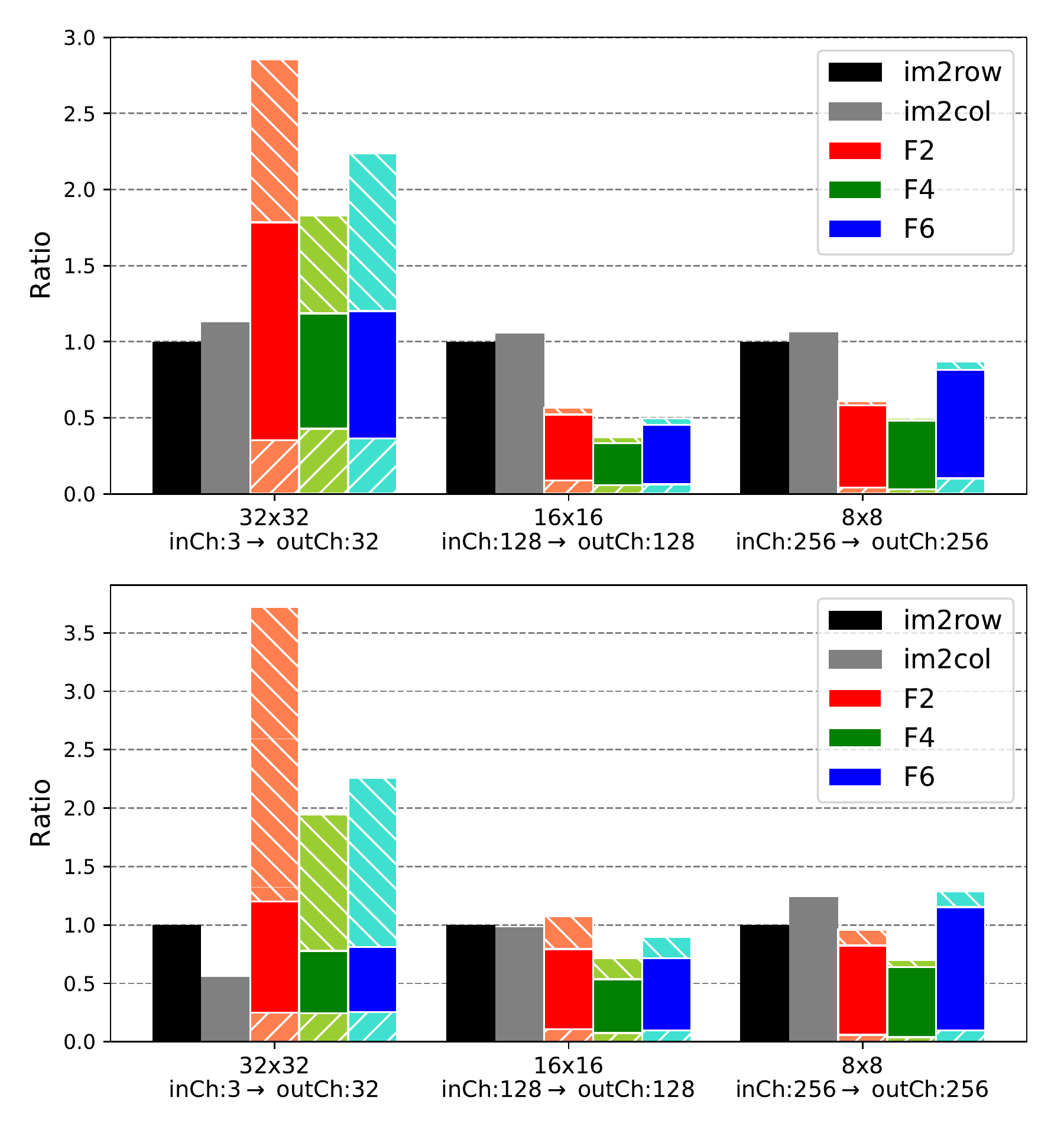}
    \end{center}
    \vspace{-7mm}
    \caption{Latencies (normalized \textit{w.r.t} \textit{im2row}) to execute different layers of the ResNet-18 network measured in A73 (above) and A53 (below). For Winograds, solid colour bar regions represent the element-wise GEMM stage, below and above it are respectively the input and output transformation costs. Measured for FP32 and with default Winograd transforms. }
    \label{fig:ratiosTransforms}
    \vspace{-5mm}
\end{figure}

Figure \ref{fig:latency} shows a small portion of the obtained latency measurements for our benchmark in FP32. An almost identical pattern appears when using 8-bit arithmetic. In Figure \ref{fig:ratiosTransforms} we show the speedups that Winograd convolutions offer at different layers of a ResNet-18 network. Our observations can be summarized in three points:

\textbf{Input layers do not benefit from Winograd.}  This is primarily because the matrices in the element-wise GEMM stage are not large enough to compensate for the costs of transforming the input and output patches (see Figure \ref{benchmarks} and \ref{fig:ratiosTransforms}). They represent a significant portion (up to 65\% and 75\% respectively on the A73 and A53) of the total costs for convolving a RGB $32\times32$ input expanded to 32 channels. 
Similar ratios can be observed for other input sizes. In spite of this, this first layer accounts for a marginal portion of the total latency of the model, often below 1ms. 
    
\textbf{Optimal $\mathbf{m}$ is a function of input width and height.}  For an input with sufficient number of channels, e.g. 32 channels and up, we observe a consistent pattern alternating between $F4$ and $F6$ as the channel dimension of the output increase. This alternation comes as a result of the impossibility of subdividing the input into an integer number of $(m+r-1)\times(m+r-1)$ patches, and therefore having to waste calculations when operating around the matrix edges. This pattern is invariant to different $inCh\rightarrow outCh$ configurations and fades away as input dimensions exceed $40\times40$, where F6 consistently becomes the fastest.

\textbf{Winograd transforms are costly.}  Excluding inputs with very few channels, the cost of performing the transformations to/from the Winograd domain can exceed 25\% of the overall costs. These costs become negligible as the input width and height decrease, but the rate at which this happens also depends on the hardware. Our Winograd-aware pipeline formulation doesn't impose any constrains on how the transforms are learnt. This results in dense transforms (as opposed to the default transforms witch contain zeros) and therefore applying them require additional compute. Table \ref{tab:ResNetCIFARcomp} includes this latency overhead in models making use of learned transforms. In Appendix \ref{overheadTransforms} we provide more details on how dense transforms impact overall latency.

On A53, the speedups from FP32 Winograd convolutions are smaller than On A73. We argue this comes as a results of the differences in the memory subsystem, limiting the lower-end CPU to efficiently operate with larger tensors. These speedups grow significantly when leveraging INT8 arithmetic, made possible by winograd-aware training. Concretely, INT8 Winograd increases the speedup on the A53 by a factor of almost $1.5\times$ compared to Winograd in FP32, as shown in WA\textsubscript{F4} configurations in Table \ref{tab:ResNetCIFARcomp} -- at the cost of 1.1\% accuracy in CIFAR-10. In the case of the more challenging CIFAR-100 dataset, the drop in accuracy is more severe. However, our WA\textsubscript{F2} layers offer attractive speedups for INT8 with no drop in accuracy. We rely on wiNAS to minimize this degradation with small impact on latency. 

\subsection{wiNAS Networks}

\begin{table*}[t]
    \centering
    \small
    \begin{tabular}{l c c c c c c c}
        \toprule
        Conv. & Bits & \multicolumn{2}{c}{Accuracy (\%)} & \multicolumn{2}{c}{Cortex-A53} & \multicolumn{2}{c}{Cotex-A73} \\
        
        Type & act. / param. & CIFAR-10 & CIFAR-100 & Latency (ms) & Speedup & Latency (ms) & Speedup \\
        \midrule
        \textit{im2row} & \multirow{7}{*}{32 / 32}  & 93.16 & 74.62 & 118 & - & 85 & - \\
        \textit{im2col} &  & 93.16 & 74.62  & 156 & $0.76\times$ & 102 & $0.83\times$ \\
        W\textsubscript{F2} &   & 93.16 & 74.60 & 126 & $0.94\times$ & 56  &  $1.52\times$ \\
        W\textsubscript{F4} &   & 93.14 & 74.53  & 97 & $1.22\times$ & 46  & $1.85\times$ \\
        WA\textsubscript{F2}* &   & 93.46 & 74.69  & 126 & $0.94\times$ & 56 & $1.52\times$ \\
        WA\textsubscript{F4} &   & 93.54 & 74.98  & $122^{\dagger}$ & $0.92\times$ & $54^{\dagger}$  & $1.58\times$ \\
        wiNAS\textsubscript{WA} &  & 93.35  & 74.71 & $123^{\dagger}$ & $0.96\times$ & $56^{\dagger}$  & $1.52\times$ \\
        \midrule
        \textit{im2row} & \multirow{5}{*}{8 / 8} & 93.20 & 74.11 & 117 & $1.01\times$ & 54 & $1.57\times$\\
       \textit{im2col} &  & 93.20 & 74.11 & 124 & $0.95\times$ & 59 & $1.45\times$ \\
        WA\textsubscript{F2}* &  & 93.72 & 73.71 & 91 & $1.30\times$ & 38 & $2.24\times$ \\
        WA\textsubscript{F4} &  & 92.46 & 72.38 & $82^{\dagger}$ & $\bm{1.44\times}$ & $35^{\dagger}$  & $\bm{2.43\times}$\\
        wiNAS\textsubscript{WA} &  & 92.71 & 73.42 & $88^{\dagger}$ / $91^{\dagger}$ & $1.34\times$ / $1.30\times$ & $35^{\dagger}$ / $36^{\dagger}$ & $2.43\times$ / $2.36\times$ \\
        \midrule
        wiNAS\textsubscript{WA-Q} & auto & 92.89 & 73.88 & $74^{\dagger}$ / $97^{\dagger}$  & $\bm{1.60\times}$ / $1.22\times$ & $32^{\dagger}$ / $43^{\dagger}$ & $\bm{2.66\times}$ / $1.98\times$\\
        \bottomrule
    \end{tabular}
    \caption{Performance in terms of accuracy and latency (ms) of ResNet-18 when convolutions are implemented with different algorithms and for different quantization levels. We show that Winograd-aware (WA\textsubscript{F2/4}) layers combine the speedups of Winograd convolutions with those of INT8 arithmetic, with little to no accuracy loss in some cases. This is not possible with existing Winograd (W\textsubscript{F2/4}) formulations. Latency is measured on Arm Cortex-A73 and A53 cores. For the last two rows, wiNAS found different optimizations for each dataset. We show latencies for CIFAR-10 on the left and CIFAR-100 on the right. Speedups are shown against \textit{im2row} in FP32. (*) With default Winograd transforms. ($\dagger$) Includes worst case latency increase due to be using learned transform, which are often dense.}
    \label{tab:ResNetCIFARcomp}
\end{table*}

Choosing the convolution algorithm that minimizes overall network latency can be easily done by looking at the benchmark results. However, since the accuracy of Winograd convolutions degrade rapidly in reduced precision networks, selecting the fastest algorithm for each layer without sacrificing accuracy, is not straight forward. 

When using wiNAS\textsubscript{WA}, values of $\lambda_{2}$ larger than 0.05 consistently result in models with the same layer configuration as those in  WA\textsubscript{F4} (described in section \ref{experimentalsetup_WA_networks}). When lowering the impact of latency in Eq.~\ref{eq:arch_search_loss} loss function, we observed several $F4$ Winograd-aware layers were replaced with either \textit{im2row} or $F2$, at the cost of less than 9 ms latency increase in the worst case, an INT8 model on the A53 for CIFAR-100. These models resulted in similar accuracies in FP32 and reached $0.32\%$ and $1.1\%$ higher accuracies in INT8 for CIFAR-10 and CIFAR-100 respectively. Despite CIFAR-100 models sacrificing more latency in order to recover accuracy, they remained faster than WA\textsubscript{F2} at INT8.

When introducing quantization in the search performed by wiNAS\textsubscript{WA-Q}, the accuracy gap is almost closed for CIFAR-10 and further reduced for CIFAR-100. This comes primarily as a result of relying on higher bit-widths for the first layers in the network. In both cases, we maintain attractive speedups compared to \textit{im2row} and Winograd convolutions in FP32, specially on the A73. All the ResNet-18 architectures optimized with wiNAS are described in \ref{winasArch}.

\section{Discussion}
\label{discussion}

In this section we present some of the challenges of training Winograd-aware networks and propose lines of future work.

A direct implementation of Eq. \ref{eq:winogradmatrix} requires saving the intermediate outputs of each matrix-matrix multiplication, since these are needed for back-propagation. This results in high memory usage. In this work we had to rely on \textit{gradient checkpointing}~\cite{gradientCheckpointing} to lower the memory peak during training, at the cost of additional computation. We believe a native CUDA implementation of the Winograd-aware layers with better memory reuse would ease this problem.
    
Learning larger models (with width multipliers 0.75 and 1.0) proved challenging for $F4$ and $F6$ when introducing quantization. Using other types of quantization would likely help. In particular per-channel affine quantization, as in \citet{Jacob_2018}. Also, enabling different bit-widths throughout Eq. \ref{eq:winogradmatrix} could help mitigate the accuracy drop. 
    
It is known that bad polynomial points for constructing $G$, $B^\top$ and $A^\top$ introduce significant deviations in the result of computing Winograd convolutions compared to that of normal convolutions. We observed that good starting points are also important even when learning the Winograd transformations. Polynomial points specifically tailored for quantized Winograd could alleviate some of the degradation that we observed with increased tile size. 
    
In this work we focused on mobile CPUs, but we expect these benefits to be also applicable to GPUs. However, to further maximize the speedups that Winograd-aware layers for quantized CNNs offer, a custom hardware implementation in the form of an accelerator would be preferable.

\section{Conclusion}
\label{conclusion}

Running CNN-based applications that require standard convolutional layers is challenging in compute-constrained devices such as mobile CPUs. This paper presents Winograd-aware layers as the building block to combine the benefits of quantized networks and fast Winograd convolutions. We studied Winograd-aware layers with different tile sizes, three quantization levels and on three popular datasets. We found that allowing the transformation matrices to evolve during training resulted in significantly better models. With wiNAS we leveraged Winograd-aware layers and latency metrics from off-the-shelf mobile CPUs and found architectures that helped minize the numerical instability of Winograd. A Winograd-aware ResNet-18 quantized to INT8 offers up to $1.32\times$ faster inference for only a marginal accuracy drop compared to existing Winograd implementations, which are limited to FP32. This network is also $1.54\times$ faster than an optimized \textit{im2row} implementation using INT8 arithmetic.

\small
\bibliography{references}
\bibliographystyle{mlsys2020}

\cleardoublepage
\normalsize
\appendix
\section{Appendix}
\label{appendixExtra}

\subsection{Winograd-aware layers for other architectures}
\label{otherArchitectures}

The results of our study of Winograd-aware networks presented Section \ref{experimentalresults} showed multiple configurations of the ResNet-18 architecture at different width-multipliers, bit-widths, quantization levels and convolution algorithms. Here, we present a similar analysis for two other popular architectures for image classification. We limit our study to the full models (i.e. mult=1.0) We show results for SqueezeNet~\cite{i2016squeezenet} in Table \ref{tab:SqueezeNetresults} and for ResNeXt~\cite{Xie2016AggregatedRT} in Table \ref{tab:ResNeXtresults}. These results align with what was observed for ResNet-18: In the presence of quantization, learning the Winograd transformations (\textit{flex} configurations) resulted in superior performance than using the default (\textit{static}) transformations. All experiments used the same hyper-parameters as described in Section \ref{experimentalsetup}.

\begin{table}[ht]
    \centering
    \small
    \begin{tabular}{l c c c c}
        \toprule
        Conv. & Bits & WA & \multicolumn{2}{c}{Accuracy (\%)} \\
        Type & act. / param. & trans. & CIFAR-10 & CIFAR-100 \\
        \midrule
        \textit{im2row} & \multirow{5}{*}{32 / 32}  & - & 91.13 & 69.06 \\
        WA\textsubscript{F2} &  & static & 91.31 & 69.42 \\
        WA\textsubscript{F2} &  & flex & 91.25 & 69.36 \\
        WA\textsubscript{F4} &  & static & 91.23 & 69.14 \\
        WA\textsubscript{F4} &  & flex & 91.41 & 69.32 \\
        \midrule
        \textit{im2row} & \multirow{5}{*}{8 / 8} & - & 91.15 & 69.34 \\
        WA\textsubscript{F2} &  & static & 90.88 & 70.06 \\
        WA\textsubscript{F2} &  & flex & 91.03 & 70.18 \\
        WA\textsubscript{F4} &  & static & 79.28 & 55.84 \\
        WA\textsubscript{F4} &  & flex & 90.72 & 69.73 \\
        \bottomrule
    \end{tabular}
    \caption{Comparison between standard convolutions (\textit{im2row}) and Winograd-aware layers for SqueezeNet. With INT8 quantization and using the default transformation matrices (\textit{static}), larger tile sizes (i.e. F4) introduce substantial numerical error and result in a sever accuracy drop. This drop in accuracy is significantly reduced if the transformations are learnt (\textit{flex}).}
    \label{tab:SqueezeNetresults}
\end{table}

\begin{table}[ht]
    \centering
    \small
    \begin{tabular}{l c c c c}
        \toprule
        Conv. & Bits & WA & \multicolumn{2}{c}{Accuracy (\%)} \\
        type & act. / param. & trans. & CIFAR-10 & CIFAR-100 \\
        \midrule
        \textit{im2row} & \multirow{5}{*}{32 / 32}  & - & 93.17 & 74.54 \\
        WA\textsubscript{F2} &  & static & 93.19 & 74.66 \\
        WA\textsubscript{F2} &  & flex & 93.08 & 74.58 \\
        WA\textsubscript{F4} &  & static & 93.24 & 74.47 \\
        WA\textsubscript{F4} &  & flex & 93.15 & 74.62 \\
        \midrule
        \textit{im2row} & \multirow{5}{*}{8 / 8}  & - & 93.40 & 74.89 \\
        WA\textsubscript{F2} &  & static & 92.93 & 75.32 \\
        WA\textsubscript{F2} &  & flex & 93.11 & 75.80 \\
        WA\textsubscript{F4} &  & static & 76.73 & 51.20\\
        WA\textsubscript{F4} &  & flex & 93.29 & 75.35 \\
        \bottomrule
    \end{tabular}
    \caption{Comparison between standard convolutions (\textit{im2row}) and Winograd-aware layers for ResNeXt\textunderscore20($8\times16$). With INT8 quantization and using the default transformation matrices (\textit{static}), larger tile sizes (F4) introduce substantial numerical error and result in a sever accuracy drop. This drop in accuracy is significantly reduced if the transformation matrices are learnt (\textit{flex}).} 
    \label{tab:ResNeXtresults}
\end{table}

For both architectures, INT8 Winograd-aware $F4$ models with learnt Winograd transformations did not result in a accuracy gaps as pronounced as the ones reported for ResNet-18 in Section \ref{experimentalresults}. These models even surpass the \textit{im2row} baselines for CIFAR-100. We argue this is because SqueezeNet and ResNeXt\textunderscore($8\times16$) have fewer $3\times3$ convolutional layers (8 and 6, respecitvely) compared to ResNet-18, which has 16. Therefore, the succession of fewer convolutional layers implemented as Winograd convolutions reduces the overall impact of numerical error.  


\begin{figure*}[t]
    \centering
    \includegraphics[width=0.7\linewidth]{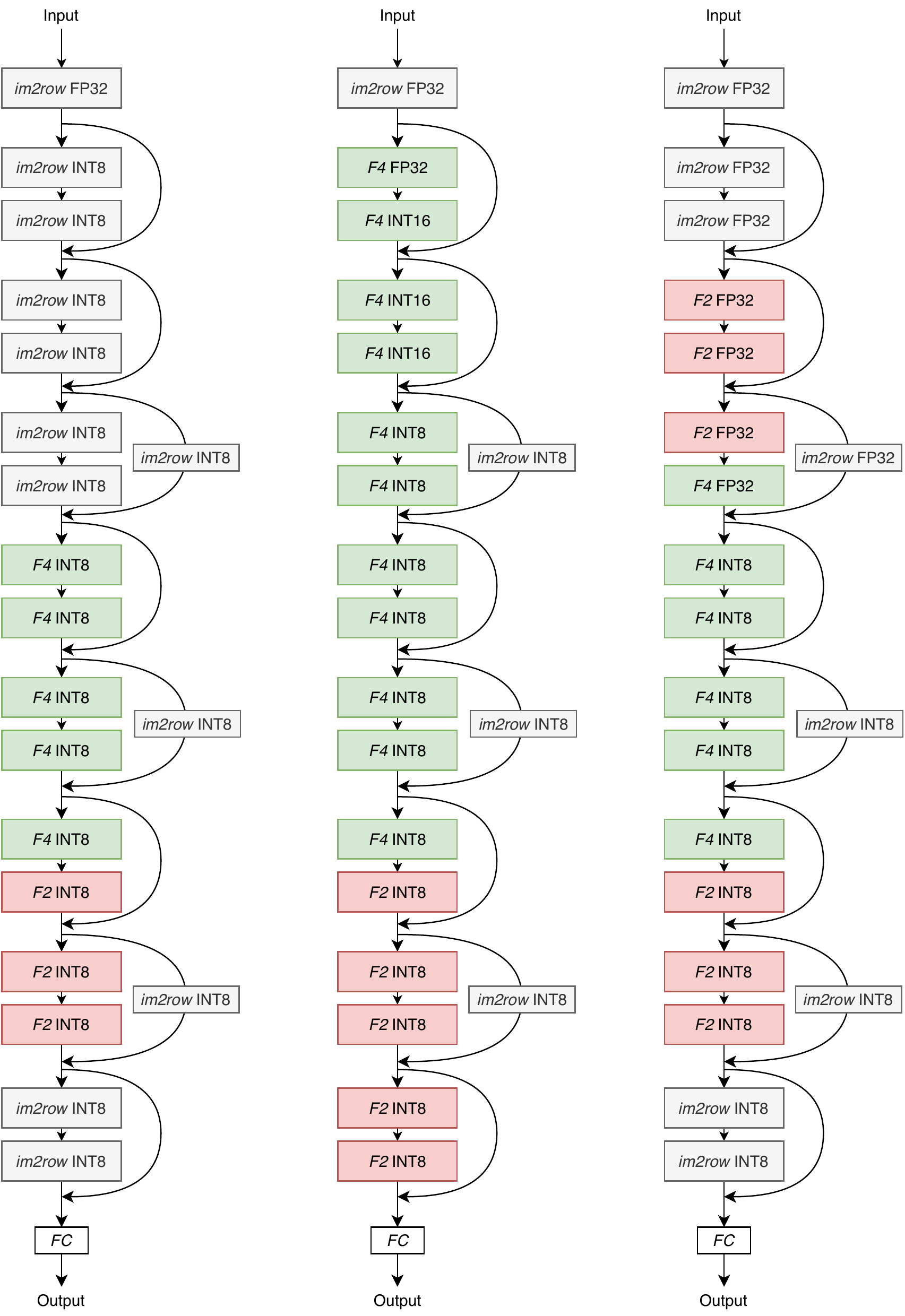}
    \caption{Resulting architectures after optimizing a ResNet-18 macro-architecture using wiNAS. For wiNAS\textsubscript{WA} and CIFAR-100, the architecture resulted is shown on the left. With wiNAS\textsubscript{WA-Q}, that introduces quantization in the search space, the optimization resulted in different architectures for CIFAR-10 (middle) and CIFAR-100 (right), evidencing the difference in complexity of the latter.}
    \label{fig:winasarch}
\end{figure*}

\subsection{Overhead of Learnt Winograd Transforms}
\label{overheadTransforms}

The default Winograd transformation matrices contain varying amounts of 0’s. For $F2$ the sparsity ratios are 50\%, 33\% and 25\% respectively for $B^{T}$, $G$ and $A^{T}$. From the construction process of these matrices and specially the choice of \textit{polynomial points}, we would expect lower sparsity ratios as these transforms are adjusted for larger input patches. For example, for the default transforms $F4$ these ratios are 22\%,  22\% and 25\%. For implementations of matrix-matrix multiplications that can exploit data sparsity, as is the case of Arm's Compute Library, having zeros means less compute which often translate into lower latencies.

The Winograd-aware formulation here presented doesn't impose restrictions on how the learnt transform should look like. As a consequence, the resulting transforms rarely contain zeros. This translates in additional compute for input $B^{T}dB$ and output $A^{T}yA$ transforms. The impact of using dense, learnt, transforms for WA\textsubscript{F4} models running on a Cortex-A73 is a latency increase of ~17\% (+8ms) and ~20\% (+6ms) for FP32 and INT8 respectively for a ResNet18 network. This increase in latency is higher on the Cortex-A53 since the Winograd transforms are proportionally more expensive on this core. These penalties represent the worst case performance increase, assuming the transforms are compute bound. However, we believe that due to the access patterns of the Winograd transform kernels (gather and scatter across a wide area of memory) at least some of the performance of the transforms results from misses in the cache hierarchy and so some additional computation can be tolerated without necessarily increasing execution time. 

We note that the impact for $F2$ models is considerably higher especially since the original transforms G and A are, not only sparse, but binary and the learnt ones are not. However, these penalties are never met in practice since $F2$ Winograd-aware models with default transforms can perform equally well as those with learnt transforms (as shown in Figure \ref{fig:flexvsstatic} and Tables \ref{tab:SqueezeNetresults} and \ref{tab:ResNeXtresults}) even in INT8.

Even with the performance loss due to the learnt transforms, we're still demonstrating some (non-negligible) $1.54\times$ and $1.43\times$ speedup compared to INT8 \textit{im2row} for A73 and A53 respectively. To the best of our knowledge this is the first time INT8 Winograd convolutions are empirically proven to work.

\subsection{Architectures optimized with wiNAS}
\label{winasArch}

Our framework wiNAS, takes a given macro-architecture and optimizes each $3\times3$ convolutional layer by choosing from direct convolution or different Winograd configurations. For the search, all $1\times1$ convolutions were fixed to use \textit{im2row}.

For wiNAS\textsubscript{WA} in FP32, the resulting architecture only substituted the last convolution layer with \textit{im2row} instead of $F2$. The rest of the layers remained unchanged from the WA\textsubscript{F4} configuration (which was described in Section \ref{experimentalsetup_WA_networks}). The same micro-architecture was used in CIFAR-10 and CIFAR-100.

For wiNAS\textsubscript{WA} with 8-bit quantization and CIFAR-10, wiNAS replaced the 5\textsuperscript{th} and second last convolutional layers with \textit{im2row}, instead of $F4$ and $F2$ respectively. For CIFAR-100, more optimization was compared to WA\textsubscript{F4}. The resulting micro-architecture optimization is shown in Figure \ref{fig:winasarch} (left).

When introducing quantization in the search space, wiNAS\textsubscript{WA-Q}, the resulting architectures are shown in Figure \ref{fig:winasarch} for both CIFAR-10 (middle) and CIFAR-100 (right).

\end{document}